\documentclass[10pt,twocolumn,letterpaper]{article}

\usepackage[pagenumbers]{style}

\definecolor{darkred}{rgb}{0.7, 0.0, 0.0}

\usepackage{bbm}
\usepackage{multirow}
\usepackage{booktabs}

\definecolor{blue}{rgb}{0.21,0.49,0.74}
\usepackage[pagebackref,breaklinks,colorlinks,allcolors=blue]{hyperref}

\usepackage{cuted}
\providecommand{\methodabbr}{NVComposer\xspace}
\providecommand{\methodfullname}{Novel View Composer\xspace}
\providecommand{\methodfullnamebold}{\textbf{N}ovel \textbf{V}iew \textbf{Composer}\xspace}

\newcommand{\blfootnote}[1]{\begingroup
	\renewcommand\thefootnote{}\footnote{#1}\addtocounter{footnote}{-1}
	\endgroup}
\title{\methodabbr: Boosting Generative Novel View Synthesis with \\ Multiple Sparse and Unposed Images}

\author{
Lingen Li$^{1, 2}$ \quad Zhaoyang Zhang$^{2\dagger}$ \quad Yaowei Li$^{2, 3}$ \quad Jiale Xu$^{2}$ \quad Wenbo Hu$^{2}$ \quad Xiaoyu Li$^{2}$  \\ \quad Weihao Cheng$^{2}$ \quad Jinwei Gu$^{1}$ \quad Tianfan Xue$^{1}$  \quad Ying Shan$^{2}$\\
{
$^{1}$The Chinese University of Hong Kong \ 
$^{2}$ARC Lab, Tencent PCG \
$^{3}$Peking University}
\vspace{-30pt}
}

\begin{document}
\maketitle
\blfootnote{$\dagger$ Project Lead.}
\blfootnote{Project Page: \href{https://lg-li.github.io/project/nvcomposer}{https://lg-li.github.io/project/nvcomposer} }
\begin{strip}
\centering
\vspace{-30pt}
\includegraphics[width=1\linewidth]{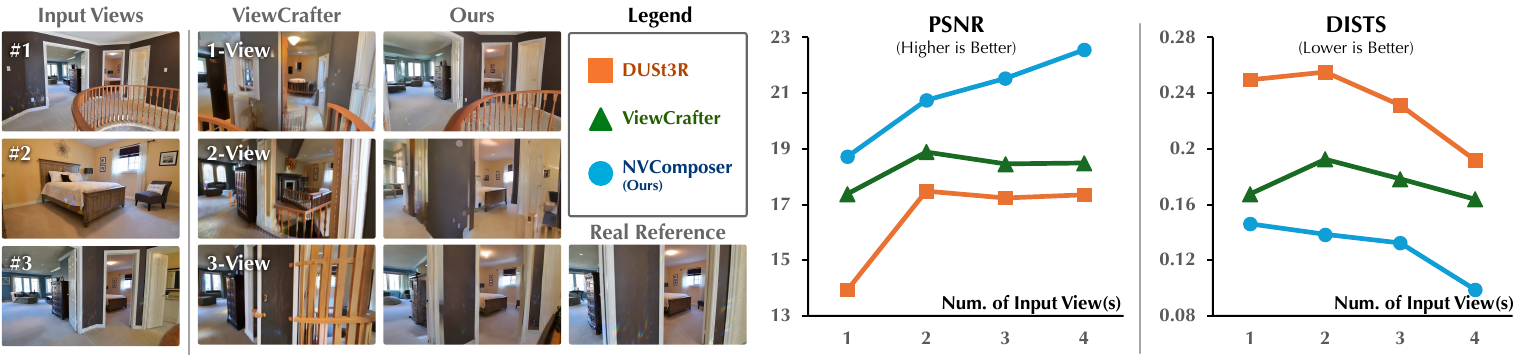}
\vspace{-20pt}
    \captionof{figure}{
    As the number of unposed input views increases, \methodabbr (blue circle) effectively uses the extra information to improve NVS quality. In contrast, ViewCrafter~\cite{yu2024viewcrafter} (green triangle), which relies on external multi-view alignment (via pre-reconstruction from DUSt3R~\cite{wang2024dust3r}), suffers performance degradation as the number of views grows due to instability of the external alignment. This result contradicts the common expectation that ``more views lead to better performance." Please refer to \cref{sec:eval-res} for full results.
    }
    \label{fig:teaser}
\end{strip}
\vspace{-20pt}
\providecommand{\methodabbr}{NVComposer\xspace}
\providecommand{\methodfullname}{Novel View Composer\xspace}
\providecommand{\methodfullnamebold}{\textbf{N}ovel \textbf{V}iew \textbf{Composer}\xspace}

\begin{abstract}

\vspace{10pt}
Recent advancements in generative models have significantly improved novel view synthesis (NVS) from multi-view data. However, existing methods depend on external multi-view alignment processes, such as explicit pose estimation or pre-reconstruction, which limits their flexibility and accessibility, especially when alignment is unstable due to insufficient overlap or occlusions between views.
In this paper, we propose \methodabbr, a novel approach that eliminates the need for explicit external alignment. \methodabbr enables the generative model to implicitly infer spatial and geometric relationships between multiple conditional views by introducing two key components: 1) an image-pose dual-stream diffusion model that simultaneously generates target novel views and condition camera poses, and 2) a geometry-aware feature alignment module that distills geometric priors from dense stereo models during training.
Extensive experiments demonstrate that \methodabbr achieves state-of-the-art performance in generative multi-view NVS tasks, removing the reliance on external alignment and thus improving model accessibility. Our approach shows substantial improvements in synthesis quality as the number of unposed input views increases, highlighting its potential for more flexible and accessible generative NVS systems.
\vspace{-15pt}
\end{abstract}
    
\section{Introduction}
\label{sec:intro}
\providecommand{\methodabbr}{NVComposer\xspace}
\providecommand{\methodfullname}{Novel View Composer\xspace}
\providecommand{\methodfullnamebold}{\textbf{N}ovel \textbf{V}iew \textbf{Composer}\xspace}

\begin{figure*}[htbp]
  \centering
    \includegraphics[width=1\linewidth]{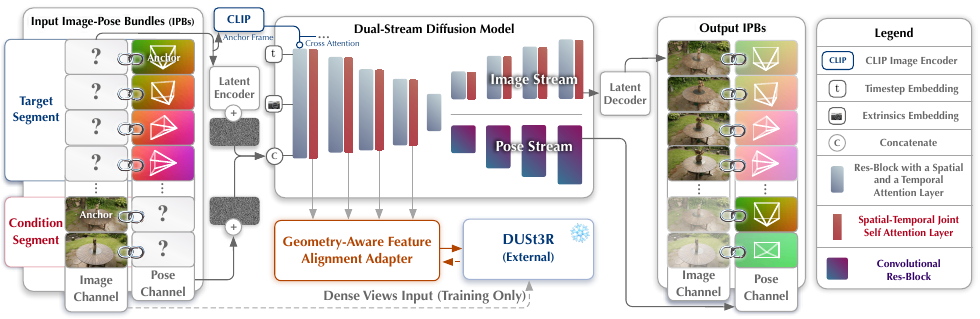}
    \vspace{-20pt}
    \caption{Framework illustration of \methodabbr. It contains an \textbf{image-pose dual-stream diffusion model} that generates novel views while implicitly estimating camera poses for conditional images, and a \textbf{geometry-aware feature alignment adapter} that uses geometric priors distilled from pretrained dense stereo models~\cite{wang2024dust3r}.
    }
    \label{fig:main-framework}
\vspace{-10pt}
\end{figure*}

With recent advances in generative models, generative novel view synthesis (NVS) methods have drawn considerable attention~\cite{liu2023zero,shi2023zero123++,long2024wonder3d,wu2024reconfusion,gao2024cat3d,yu2024viewcrafter} due to its ability to synthesize novel views with only one or a few images. 
Unlike reconstruction-based NVS methods, where dense-view images with a full coverage of the scene are required~\cite{mildenhall2021nerf, kerbl20233d, wang2024dust3r,leroy2024grounding}, generative NVS methods could take only one or a few views as inputs, completing unseen parts of a scene with plausible content~\cite{rockwell2021pixelsynth, yu2021pixelnerf}. This capability is particularly useful in applications where capturing extensive views is impractical, offering greater flexibility and efficiency for virtual scene exploration and content creation.

In addition to generate novel views from a single input image, generative NVS methods~\cite{yu2024viewcrafter,liu2024reconx,gao2024cat3d} have demonstrated more flexible utility by reducing ambiguity through giving additional input images~\cite{xu2025sparp}. 
To leverage multi-view images in the generative NVS tasks, existing methods~\cite{gao2024cat3d, yu2024viewcrafter, wu2024reconfusion} all rely on external multi-view alignment processes before generation. For example, assuming accurate poses of condition images are given (through \textit{explicit pose estimation})~\cite{gao2024cat3d} or generate novel views conditioned on results extracted from reconstructive NVS methods (through \textit{pre-reconstruction})~\cite{wu2024reconfusion, liu2024reconx, yu2024viewcrafter}. However, in the case of the overlap region being small and hard to do stereo matching, external multi-view alignment processes like camera pose estimation becomes unreliable~\cite{xu2025sparp}. As a result, multi-view generative NVS methods which heavily rely on the external alignment also tend to fail as shown in \cref{fig:teaser}.

To overcome this limitation, we explore the possibility of removing the dependency on the external alignment process and propose \methodfullnamebold~(\methodabbr) which could generate novel views from spare unposed images without relying on any external alignment process. Our method is able to generate reasonable results in the sparse views with small overlap and large occlusion.

Firstly, to leverage the powerful generation ability of the video diffusion model, we use a pre-trained video diffusion model as the backbone of our \methodabbr with unposed images as the condition in synthesis. Without external alignment of these images, we introduce a novel dual-stream diffusion model in \methodabbr to learn the relative poses of condition images during generation. The dual-stream diffusion model not only generates novel views but also implicitly predicts the correct pose relationships between the condition images, ensuring that the model understands the relative positioning of the condition images in the scene and uses them to synthesize novel views correctly.

Moreover, to generate more view-consistent results, we employ features produced by a pretrained dense stereo model~\cite{wang2024dust3r} to train our model with geometry awareness. Unlike previous methods~\cite{liu2024reconx, yu2024viewcrafter} that directly rely on the reconstruction results from dense stereo model as input, we propose a more flexible and accessible geometry-aware feature alignment adapter. This adapter aligns our model's features with the predicted 3D features of the dense stereo model and requires no explicit reconstruction during inference. This strategy allows us to distill 3D knowledge implicitly from the dense stereo model~\cite{wang2024dust3r}. Experiments (\cref{sec:eval-res}) demonstrate that this implicit geometry-aware learning achieves competitive performance compared to explicit reconstruction-relied methods~\cite{yu2024viewcrafter, wang2024dust3r}. It provides enhanced accessibility and flexibility, as it operates in an end-to-end manner and eliminates the need for an extra step of explicit pose estimation or pre-reconstruction during the inference.

To train our model, we construct a mixed dataset from different sources such as video~\cite{zhou2018stereo, ling2024dl3dv, reizenstein2021common} data and 3D~\cite{deitke2023objaverse} data with real indoor and outdoor scenes as well as synthetic 3D objects. \methodabbr is trained on this diverse dataset using as few as one to four randomly sampled unposed condition images. Extensive experiments demonstrate that, when provided with multiple unposed input views, \methodabbr outperforms state-of-the-art controllable video diffusion models and generative NVS models.

Our key contributions are summarized as follows:
\begin{itemize}[itemsep=-0.1em]
    \item We introduce the first pose-free multi-view generative NVS model for both scenes and objects, without the requirement for explicit multi-view alignment processes on input images. %
    \item Our proposed design which includes image-pose dual-stream diffusion and geometry-aware feature alignment adapter, highlights a promising direction for creating more flexible and accessible generative NVS systems.
    \item Our \methodabbr achieves state-of-the-art performance on generative NVS tasks for both scenes and objects when given multiple unposed input views.
\end{itemize}

\providecommand{\methodabbr}{NVComposer\xspace}
\providecommand{\methodfullname}{Novel View Composer\xspace}
\providecommand{\methodfullnamebold}{\textbf{N}ovel \textbf{V}iew \textbf{Composer}\xspace}

\section{Related Work}

\paragraph{Single-View Generative Novel View Synthesis.} Early NVS methods using feed-forward networks map a single input image to new views~\cite{tucker2020single, wang2021ibrnet}, but are limited to small rotations and translations due to the restricted information from one input image. Generative NVS method effectively hallucinate unseen views given limited input~\cite{wiles2020synsin, rockwell2021pixelsynth, chan2023generative}. Recent advances in diffusion models~\cite{ho2020denoising, song2020denoising} leverage rich image priors for NVS to synthesize more reasonable multi-view content~\cite{liu2023zero,long2024wonder3d,kant2024spad,voleti2025sv3d} by utilizing pretrained image diffusion models~\cite{rombach2022high}.
Zero-1-to-3\cite{liu2023zero} fine-tunes a latent diffusion model~\cite{rombach2022high} with image pairs and their relative poses for novel view synthesis from a single image. Wonder3D\cite{long2024wonder3d} incorporates image-normal joint training and view-wise attention to enhance generative quality. SV3D~\cite{voleti2025sv3d} fine-tunes a video diffusion model~\cite{blattmann2023stable} for NVS of synthetic objects.

However, single-view NVS models struggle to infer occluded or missing details due to the limited information from one viewpoint, making them less practical for real-world applications requiring complete scene understanding.

\vspace{-15pt}
\paragraph{Multi-View Generative Novel View Synthesis.} To overcome single-view limitations, multi-view conditioned generative NVS utilizes images from multiple viewpoints~\cite{wu2023ifusion, gao2024cat3d, yu2024viewcrafter, liu2024reconx}, enhancing the fidelity of generated views by capturing finer details and accurate spatial relationships.
iFusion~\cite{wu2023ifusion} employs a pretrained Zero-1-to-3~\cite{liu2023zero} as an inverse pose estimator and tunes a LoRA~\cite{hu2021lora} adapter for each object to support multi-view NVS. CAT3D~\cite{gao2024cat3d} uses Plücker ray embeddings~\cite{sitzmann2021light} as pose representations and masks the target view for inpainting, allowing flexibility in the number of conditioning images. ViewCrafter~\cite{yu2024viewcrafter} reconstructs an initial point cloud using a dense stereo model~\cite{wang2024dust3r} and then employs a video diffusion model to inpaint missing regions in rendered novel views.

These methods, however, rely on accurate pre-computed poses of conditional images. Sparse views that lead to inaccurate poses can significantly degrade the quality of generated views, limiting their robustness in practical scenarios.

\vspace{-15pt}
\paragraph{Video Diffusion Models.} Advancements in diffusion models have extended their capabilities from static images to dynamic videos, enabling temporally coherent video generation conditioned on various inputs~\cite{ho2022video, blattmann2023align, blattmann2023stable, xing2025dynamicrafter, esser2023structure, wang2024motionctrl, he2024cameractrl}. Ho \etal~\cite{ho2022video} first introduced diffusion models for video generation. Video LDM~\cite{blattmann2023align} operates in the latent space~\cite{rombach2022high} to reduce computational demands. Subsequent works enhance controllability by incorporating additional conditions. AnimateDiff~\cite{guo2023animatediff} extends text-to-image diffusion models to video by attaching motion modules while keeping the original model frozen. DynamiCrafter~\cite{xing2025dynamicrafter} introduces an image adapter for image-conditioned video generation. MotionCtrl~\cite{wang2024motionctrl} and CameraCtrl~\cite{he2024cameractrl} incorporate camera trajectory control using pose matrices and Plücker embeddings, respectively. ReCapture~\cite{zhang2024recapture} generates new camera trajectory views based on a given video.

Building upon the video diffusion model, our method leverages temporal coherence to synthesize unseen areas not included in the input images. Compared to previous controllable video diffusion models, our approach achieves better accuracy in camera controllability, offering robust performance for generative NVS tasks.

\providecommand{\methodabbr}{NVComposer\xspace}
\providecommand{\methodfullname}{Novel View Composer\xspace}
\providecommand{\methodfullnamebold}{\textbf{N}ovel \textbf{V}iew \textbf{Composer}\xspace}

\section{Methodology}
The objective of \methodabbr is to develop a model capable of generating novel views at specified target camera poses, using multiple unposed conditional images without requiring external multi-view alignment (\emph{e.g.}, explicit pose estimation).
To achieve this, we propose to enable the model itself to infer the spatial relationships of the conditional views during generation. We introduce this capability through two key strategies: (1) instead of explicitly solving for camera poses, we model pose estimation as a generative task that jointly happens with the image generation, and (2) we distill effective geometric knowledge from expert models into our generative model.

This leads to two main components of our \methodabbr, as shown in \cref{fig:main-framework}: an \textbf{image-pose dual-stream diffusion} model that generates novel target views while implicitly estimating camera poses for conditional images, and a \textbf{geometry-aware feature alignment} adapter that uses geometric priors distilled from pretrained dense stereo models~\cite{wang2024dust3r}. The design and implementation of these components are detailed below.

\subsection{Image-Pose Dual-Stream Diffusion}

Assume the model accepts $T$ elements as input and produces $T$ elements as output, where each element corresponds to an image captured within the current scene, accompanied by its pose annotation. We refer to these elements as \textit{image-pose bundles}. We partition these bundles into two segments: the first $N$ bundles constitute the \textit{target segment}, and the remaining $M$ bundles form the \textit{condition segment}, as illustrated in \cref{fig:main-framework}.

\paragraph{Image-Pose Bundles.} Specifically, let $I_t \in \mathbb{R}^{3 \times H \times W}$ denote the $t$-th RGB image, and $P_t \in \mathbb{R}^{6 \times H \times W}$ be the corresponding Plücker ray embedding~\cite{sitzmann2021light} representing the camera pose.
The conditional input and generated output are sequences of $T$ image-pose bundles for a specific scene, denoted as $\mathcal{B} = \{[I_{t}', P_{t}']\}_{t=1}^T$, where the $t$-th image-pose bundle consists of the concatenation (along the channel dimension) of the latent image $I_{t}' = \mathcal{E}(I_t) \in \mathbb{R}^{4\times \frac{H}{8} \times \frac{W}{8}}$ and the resized Plücker ray $P_t' \in \mathbb{R}^{6 \times \frac{H}{8} \times \frac{W}{8}}$. Here, $[\cdot]$ denotes concatenation and $\mathcal{E}$ represents the VAE encoder of latent diffusion.

The main difference between the target output $\mathcal{B}$ and the conditional input $\mathcal{B}{c}$ is that $\mathcal{B}$ is complete, while $\mathcal{B}{c}$ is partially masked. Specifically, for $\mathcal{B}_{c}$, image latents in the target segment (\emph{i.e.}, the first $N$ elements) and pose embeddings in the condition segment are set to zero.  
Additionally, we utilize relative camera poses in our model and designate the first elements of both the target and condition segments as the \textit{anchor view} for this relative coordinate system. This implies that these two elements are expected to have the same image content, and their camera extrinsic matrices are identity transformations. The image for the anchor view is always provided during training and inference, corresponding to the scenario where at least one conditional image is available.
With this design, our image-pose dual-stream diffusion model accepts $M$ unposed conditional images and $N$ target poses as input, and output $N$ novel view images at target poses along with $M$ predicted poses for conditional images, where $N, M \geq 1 $ and $N + M = T$. 

\paragraph{Video Prior.}
Video priors from video diffusion model has been validated to be useful for generative NVS tasks~\cite{gao2024cat3d, yu2024viewcrafter, liu2024reconx}.
To fully leverage the generative priors in \methodabbr, we initialize the dual-stream diffusion model using the pretrained weights of the video diffusion model DynamiCrafter~\cite{xing2025dynamicrafter}. We omit the frame rate and text conditions from the original video diffusion model, focusing solely on the relevant components for our task. We retain the image CLIP~\cite{radford2021learning} feature conditioning with its Q-Former-like~\cite{li2023blip} image adapter in the cross-attention layers, using the anchor view as the conditioning image. Additionally, we enhance the model's capacity to capture cross-view correspondences at different spatial locations by adding extra spatio-temporal self-attention layers after each Res-Block of original video diffusion model. 
Although camera pose information is provided in $\mathcal{B}_{c}$ as Plücker ray embeddings, we further incorporate the sequence of target camera extrinsics embedding encoded by a learnable multilayer perceptron layer from the corresponding $3 \times 4$ camera-to-world matrices $R_{c} \in \mathbb{R}^{T\times 3 \times 4}$ of all image-pose bundles, where the last $M$ elements on the temporal dimension are masked with zeros like what we have done with $\mathcal{B}_{c}$. These embeddings are added to the time-step embedding and serve as supplemental signals indicating the poses for each image-pose bundle in the target segment.

\paragraph{Pose Decoding Head Separation.} We observed that training the model to generate image-pose bundles directly is hard to converge. This is caused by the difference of the two modality: images latents $I_{t}'$ contain complex latent features representing diverse content, while the pose embeddings $P_{t}'$ are dominated by low-frequency component. This disparity can lead to interference when jointly denoising $[I_{t}', P_{t}']$ using tightly coupled network layers.

To address this issue, we design an additional decoding head specifically for denoising $P_{t}'$. As shown in \cref{fig:main-framework}, this \textit{pose decoding head} operates in parallel with the original decoding part of the diffusion denoising U-Net and follows a fully convolutional architecture similar to the original decoder. It takes as input the bottleneck features, residual connections from the encoding part, and the denoising time-step embeddings. Since the Plücker ray embeddings of poses are predominantly low-frequency and relatively straightforward to denoise, we empirically reduce the base channel number of the pose decoding head to one-tenth of that of the original decoder for images and remove its attention layers. The outputs of the pose decoding head are concatenated along the channel dimension with those of the original decoding part of the diffusion U-Net to form the final output.

\begin{figure}[t]
  \centering
    \includegraphics[width=1\linewidth]{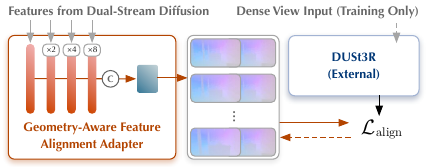}
    \caption{Structure of the geometry-aware feature alignment adapter in~\methodabbr, which aligns the internal features of the dual-stream diffusion models with the 3D point maps produces by DUSt3R~\cite{wang2024dust3r} during training. Block with notation ``$\times 2$", ``$\times 4$", and ``$\times 8$" refer to bilinear upsampling on spatial dimensions. The four red bars refer to the channel-wise MLPs.}
    \label{fig:method-geometry-aware-alignment}
\vspace{-10pt}
\end{figure}

\subsection{Geometry-aware Feature Alignment}
Since the dual-stream diffusion model is initialized from a video diffusion model that is not inherently trained with geometric constraints across views, we introduce a geometry-aware feature alignment mechanism in \methodabbr. This mechanism distills effective geometric knowledge from an external model with strong geometry priors during training. Specifically, we leverage the dense stereo model DUSt3R~\cite{wang2024dust3r}, which performs well with dense views (both target and condition images), to compute $T$ pointmaps across all views ($T_t$, for $t=1 , 2 , … , T$) relative to the anchor view $T_1$.

We then align the internal features of our diffusion model with these point maps through a geometry-aware feature alignment adapter during training. Specifically, as illustrated in \cref{fig:method-geometry-aware-alignment}, the alignment adapter (the red block on the left) takes features from the encoding part of the dual-stream diffusion U-Net, immediately after each spatial-temporal self-attention layer. These features are resized to match the spatial dimensions of the image latent inputs,  $\frac{H}{8} \times \frac{W}{8}$ (the white squares in \cref{fig:method-geometry-aware-alignment}).
The resized features retain their temporal dimension of length $T$, and all operations within the geometry-aware feature alignment adapter are temporally independent. The features are then processed by channel-wise MLPs (four red bars in \cref{fig:method-geometry-aware-alignment}) to reduce them to 320 channels, followed by a convolutional residual block that outputs a 4-D tensor $F \in \mathbb{R}^{T \times 6 \times \frac{H}{8} \times \frac{W}{8}}$.

For all item $f_t, t=1, 2, .., T$ in $F$ along the temporal dimension, we minimize the mean squared error (MSE) with the concatenated point maps produced by DUSt3R~\cite{wang2024dust3r} $D$ given the $t$-th view $I_{t}$ and the anchor view $I_{1}$:
\begin{equation}
    \mathcal{L}_{\text{align}} = \frac{1}{T}\sum_{t=1}^T \Vert f_t - D(I_1, I_t)  \Vert^2_2. 
\end{equation}

\begin{table*}[t]
  \centering
  \resizebox{\linewidth}{!}{
\begin{tabular}{@{}c|c|cccc|cccc|cccc@{}}
\toprule
\multirow{3}{*}{Model} &  \multirow{3}{*}{Views} & \multicolumn{4}{c|}{Easy} & \multicolumn{4}{c|}{Medium} & \multicolumn{4}{c}{Hard} \\ 
 &   &  \multicolumn{4}{c|}{\small{($\theta_{\text{cond}} < 10$ and $\theta_{\text{target}} < 10$)}} &   \multicolumn{4}{c|}{\small{($10 \leq \theta_{\text{cond}} < 30$ and  $10 \leq \theta_{\text{target}} < 30$)}} &  \multicolumn{4}{c}{\small{($60 \leq \theta_{\text{cond}} < 120$ and  $30 \leq \theta_{\text{target}} < 60$)}} \\ 
\cmidrule(l){3-14} 
& & PSNR$\uparrow$ & SSIM$\uparrow$ & LPIPS$\downarrow$ & DISTS$\downarrow$ & PSNR$\uparrow$ & SSIM$\uparrow$ & LPIPS$\downarrow$ & DISTS$\downarrow$ & PSNR$\uparrow$ & SSIM$\uparrow$ & LPIPS$\downarrow$ & DISTS$\downarrow$ \\ \midrule

MotionCtrl~\cite{wang2024motionctrl} & 1 & 15.0741        & 0.6071         & 0.3616            & 0.0999            & 12.0674        & 0.5667         & 0.5439            & 0.1584            & 11.6381        & 0.5276         & 0.5762            & 0.1633            \\ \midrule

CameraCtrl~\cite{he2024cameractrl}                    & 1 & 13.6082        & 0.5050         & 0.4234            & 0.1458            & 11.9639        & 0.4934         & 0.5217            & 0.1957            & 11.7599        & 0.4716         & 0.5478            & 0.2021            \\ \midrule

\multirow{4}{*}{DUSt3R~\cite{wang2024dust3r}} 
& 1 & 13.9443 & 0.5582  & 0.3914 & 0.1565  & 11.4854 & 0.4520  & 0.5570  & 0.2294   & 10.9003 &  0.4029 & 0.6089   & 0.2495      \\
 & 2 & 17.4837 & 0.6148 & 0.3582 & 0.1503 & 13.3077 & 0.4886 & 0.5434 & 0.2126  & 11.5381 & 0.4003 & 0.6407  & 0.2551   \\
 & 3 & 17.2341 & 0.6097 & 0.3585 & 0.1504  & 13.2212 & 0.4978  & 0.5287  & 0.2056   & 11.9211 & 0.4387 & 0.5942 & 0.2313    \\
 & 4 & 17.3545 & 0.6193  & 0.3541 & 0.1481  & 14.6845 & 0.5534 & 0.4892 & 0.1870  & 14.2381 &  0.5280 & 0.5295   & 0.1917      \\ \midrule

\multirow{4}{*}{ViewCrafter~\cite{yu2024viewcrafter}} & 1 & 17.3750        & 0.6670         & 0.2849            & 0.1221            & 13.6015        & 0.6016         & 0.4315            & 0.1762            & 14.0781        & 0.5894         & 0.4293            & 0.1676            \\
 & 2 & 18.8906        & 0.6685         & 0.3079            & 0.1334            & 14.2891        & 0.5947         & 0.4478            & 0.1761            & 13.5859        & 0.5537         & 0.5100            & 0.1925            \\
 & 3 & 18.4531        & 0.6548         & 0.3024            & 0.1294            & 14.1172        & 0.5913         & 0.4401            & 0.1717            & 13.7031        & 0.5620         & 0.4867            & 0.1784            \\
 & 4 & 18.4844        & 0.6553         & 0.3068            & 0.1346            & 14.7421        & 0.6011         & 0.4230            & 0.1672            & 15.1875        & 0.5874         & 0.4327            & 0.1638            \\ \midrule

\multirow{4}{*}{\methodabbr (Ours)} & 1 & 18.7227        & 0.7215         & 0.2354            & 0.0996            & 15.3101        & 0.6056         & 0.3445            & 0.1516            & 15.2115        & 0.6408         & 0.4048            & 0.1462            \\
& 2 & 20.7395        & 0.7681         & 0.1781            & 0.0793            & 16.9100        & 0.6445         & 0.2742            & 0.1198            & 15.3461        & 0.6638         & 0.3789            & 0.1384            \\
& 3 & 21.5278        & 0.7981         & 0.1522            & 0.0716            & 17.7071        & 0.7418         & 0.2759            & 0.1097            & 15.3825        & 0.6822         & 0.3699            & 0.1324            \\
& 4 & 22.5519        & 0.8226         & 0.1188            & 0.0537            & 19.5346        & 0.7847         & 0.2030            & 0.0851            & 17.8181        & 0.7359         & 0.2644            & 0.0988 \\
\bottomrule
\end{tabular}
  }
  \caption{NVS evaluation with varying numbers of input views on RealEstate10K~\cite{zhou2018stereo} for controllable video models MotionCtrl~\cite{wang2024motionctrl} and CameraCtrl~\cite{he2024cameractrl}, reconstructive model DUSt3R~\cite{wang2024dust3r}, and generative models ViewCrafter~\cite{yu2024viewcrafter} and \methodabbr. $\theta_{\text{target}}$ denotes the rotation angle between the anchor view and the furthest target view, while $\theta_{\text{cond}}$ indicates the angle between the anchor view and the furthest conditional view (when multiple conditions are used). 
  }
  \label{tab:exp-main-re10k}
  \vspace{-6pt}
\end{table*}

\subsection{Training Objectives}
We train the image-pose dual-stream diffusion model $h_{\theta}$ in \methodabbr to predict noises $\epsilon$ given the uniformly sampled denoising time step $k$, the noisy version of complete image-pose bundles $\mathcal{B}^{\{k\}}$ at time step $k$, the conditional image-pose bundles $S_{cond}$, and CLIP~\cite{radford2021learning} image feature of the anchor view $\Phi(I_1)$:
\begin{equation}
    \mathcal{L}_{\text{diff}}= \mathbb{E}_{\mathcal{B}, \epsilon, \mathcal{B}_{c}, \Phi(I_0), R_c, k}[\Vert \epsilon -  h_{\theta}(\mathcal{B}^{\{k\}}, \mathcal{B}_{c}, \Phi(I_1), R_c, k) \Vert],
\end{equation}
where $\theta$ is trainable parameters of the image-pose dual-stream diffusion model, $\Phi$ is the CLIP~\cite{radford2021learning} image feature encoder. Our total loss combines the diffusion loss $\mathcal{L}_{\text{diff}}$ and the feature alignment loss $\mathcal{L}_{\text{align}}$:
\begin{equation}
    \mathcal{L}_{\text{total}} = \mathcal{L}_{\text{diff}} + \lambda\mathcal{L}_{\text{align}},
\end{equation}
where $\lambda$ is a loss re-weighting factor.

\providecommand{\methodabbr}{NVComposer\xspace}
\providecommand{\methodfullname}{Novel View Composer\xspace}
\providecommand{\methodfullnamebold}{\textbf{N}ovel \textbf{V}iew \textbf{Composer}\xspace}

\section{Experiments}
In this section, we evaluate the performance of \methodabbr on generative NVS tasks for real-world scenes and synthetic 3D objects, followed by an analysis of the model's sub-components. More results are in the supplementary.

\subsection{Training Details}
\label{sec:training-details}
We train our model on a large-scale mixed dataset built from Objaverse~\cite{deitke2023objaverse}, RealEstate10K~\cite{zhou2018stereo}, CO3D~\cite{reizenstein2021common}, and DL3DV~\cite{ling2024dl3dv}. The sequence length of image-pose bundles $T$ is set to $16$.
To get samples from video datasets (RealEstate10K~\cite{zhou2018stereo}, CO3D~\cite{reizenstein2021common}, and DL3DV~\cite{ling2024dl3dv}), we randomly select a frame interval between 1 and $\lfloor T_{max} / N \rfloor$ to sample $T$ consecutive frames, where $T_{max}$ is the total number of frames in the scene. The value of $T_{max}$ ranges from 100 to 400 depending on the data sample. We randomly sample $N$ condition views and shuffle them within the image-pose bundle.
For samples from the 3D dataset (Objaverse~\cite{deitke2023objaverse}), we render each 3D object in two versions: one with $36$ orbit views and another with $32$ random views. Target views are sampled from the orbit renderings, and condition views are sampled from the random renderings, following the procedure described above.

We firstly train the model at a resolution of $512 \times 512$ for $10,000$ steps across all datasets. Next, we tune the model on RealEstate10K and DL3DV at a resolution of $576 \times 1024$ for $20,000$ steps for a higher resolution support. We use a learning rate of $1 \times 10^{-5}$ and perform all training on a cluster with 64 NVIDIA V100 GPUs, with an effective batch size of 128. For more details, please refer to the supplementary material.

\begin{table}[t]
  \centering
  \resizebox{\linewidth}{!}{
    \begin{tabular}{@{}c|c|cccc@{}}
    \toprule
     Model & Views & PSNR$\uparrow$ & SSIM$\uparrow$ & LPIPS$\downarrow$ & DISTS$\downarrow$ \\
     \midrule
     MotionCtrl~\cite{wang2024motionctrl} & 1 & 13.4003 & 0.5539 & 0.4004  & 0.1396 \\
     \midrule
     CameraCtrl~\cite{he2024cameractrl} & 1 & 12.2995 & 0.4692 & 0.4337  & 0.1829 \\
     \midrule
    \multirow{4}{*}{DUSt3R~\cite{wang2024dust3r}} 
    & 1 & 11.7650 & 0.4652 & 0.4900 & 0.2295  \\
    & 2 & 14.6660 & 0.5158 & 0.4531 & 0.2104 \\
    & 3 &  13.9156 & 0.5010  & 0.4699 & 0.2127  \\
    & 4 & 14.8716  & 0.5193  & 0.4478 & 0.2072 \\
    \midrule
    \multirow{4}{*}{ViewCrafter~\cite{yu2024viewcrafter}} 
    & 1 & 15.5625 &  0.4932 & 0.4122  & 0.2125 \\
    & 2 & 15.6875 & 0.4775 & 0.4417 & 0.2212 \\
    & 3 & 14.8593 & 0.4670 & 0.4617  & 0.2273 \\
    & 4 & 15.0625 & 0.4712  &  0.4549 & 0.2301 \\
    \midrule
    \multirow{4}{*}{\methodabbr (Ours)} 
    & 1 & 15.3101 & 0.6056 & 0.3445 & 0.1516 \\
    & 2 & 16.9100 & 0.6445 & 0.2742 & 0.1198 \\
    & 3 & 17.3115 & 0.6687 & 0.2558 & 0.1122  \\
    & 4 & 17.9248 & 0.6958 & 0.2277 & 0.1023 \\
    \bottomrule
    \end{tabular}
   }
  \caption{NVS evaluation on DL3DV~\cite{ling2024dl3dv}. When more unposed input views are provided, our model consistently reports higher performance.}
  \label{tab:exp-main-dl3dv}
  \vspace{-15pt}
\end{table}

\begin{figure*}[t]
  \centering
    \includegraphics[width=1\linewidth]{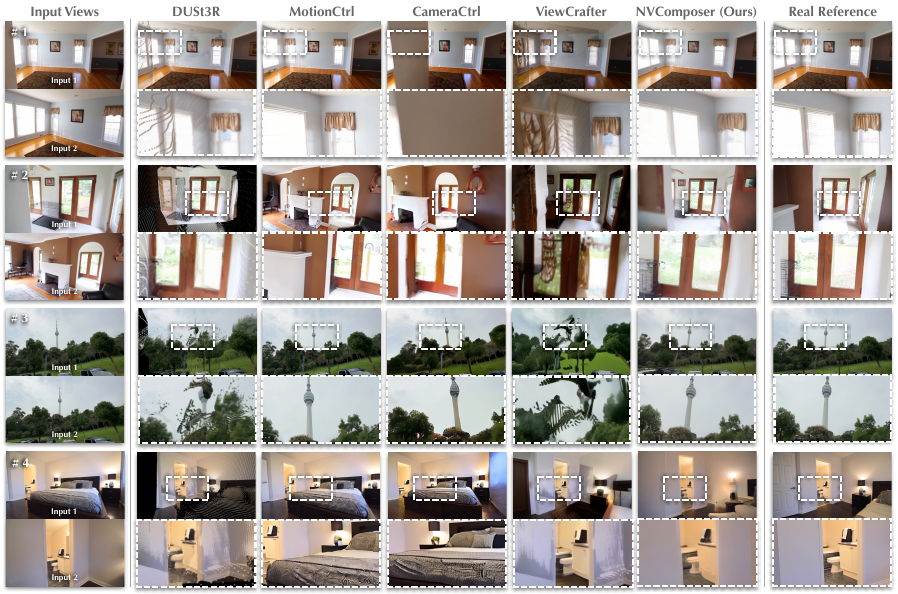}
    \caption{Visual comparison of NVS results on the RealEstate10K~\cite{zhou2018stereo} and DL3DV~\cite{ling2024dl3dv} test sets. MotionCtrl~\cite{wang2024motionctrl} and CameraCtrl~\cite{he2024cameractrl} uses the first view as input while other methods use two views as input. MotionCtrl and CameraCtrl produce incorrect camera trajectories. DUSt3R and ViewCrafter exhibit better camera control but introduce artifacts due to occlusions or misaligned multi-view inputs. Our model generates views that are visually closer to the reference. We provide zoomed-in details of the first three scenes in white boxes for a closer look. Additional visual comparisons can be found in the supplementary material.
    }
    \label{fig:eval-vis-real-scenes}
    \vspace{-10pt}
\end{figure*}

\subsection{Results}
\label{sec:eval-res}
\subsubsection{Generative NVS in Scenes}
\paragraph{Benchmark Settings.} We evaluate the performance of \methodabbr on generative NVS tasks for scenes, comparing it with four state-of-the-art models: MotionCtrl~\cite{wang2024motionctrl}, CameraCtrl~\cite{he2024cameractrl}, DUSt3R~\cite{wang2024dust3r}, and ViewCrafter~\cite{yu2024viewcrafter}.
MotionCtrl and CameraCtrl are controllable video generation models that work with a single input image, while DUSt3R is a dense stereo model for multi-view reconstructive NVS, and ViewCrafter is a multi-view generative NVS method that relies on explicit pose estimation and point cloud guidance.

For the RealEstate10K~\cite{zhou2018stereo} dataset, we categorize scenes into three difficulty levels: easy, medium, and hard. The difficulty is based on the angular distances between views, specifically the rotation angle between the anchor view and the furthest target view ($\theta_{\text{target}}$), and between the anchor and the furthest condition view ($\theta_{\text{cond}}$), when more than one condition image is used. Samples are classified as follows: (1) Easy: $\theta_{\text{cond}} < 10$ and $\theta_{\text{target}} < 10$; (2) Medium: $10 \leq \theta_{\text{cond}} < 30$ and  $10 \leq \theta_{\text{target}} < 30$; (3) Hard: $60 \leq \theta_{\text{cond}} < 120$ and  $30 \leq \theta_{\text{target}} < 60$.
We then randomly select 20 samples from the easy set, 60 from the medium set, and 20 from the hard set for evaluation.
For the DL3DV~\cite{ling2024dl3dv} dataset, we randomly select 20 test scenes.

\vspace{-10pt}
\paragraph{Results.} We measure performance by comparing generated novel views to reference images using several metrics: peak signal-to-noise ratio (PSNR), structural similarity index (SSIM)\cite{wang2004image}, and perceptual distance metrics including LPIPS\cite{zhang2018unreasonable} and DISTS~\cite{ding2020image}.
\cref{tab:exp-main-re10k} shows the numerical results on RealEstate10K and \cref{tab:exp-main-dl3dv} shows the results on the DL3DV test set. As seen, \methodabbr outperforms other methods across both datasets. \cref{fig:eval-vis-real-scenes} further demonstrates the visualized comparison among all these methods. For MotionCtrl~\cite{wang2024motionctrl} and CameraCtrl~\cite{he2024cameractrl}, pose controllability is limited. When the target camera poses involve large rotations or translations, these models generate sequences with minimal motion, failing to accurately follow the given instructions. These visual results align with the poor numerical performance observed in~\cref{tab:exp-main-re10k}.

It it noteworthy that, when there are more given input views, the performance of our method consistently increases, as we also showed in \cref{fig:teaser} before. In contrast, ViewCrafter~\cite{yu2024viewcrafter} suffers from a performance drop when the number of given views increases from one to two in the hard set. This is because the two conditional views in the hard set has large rotation difference, \emph{i.e.}, small overlapping region and possibly some occlusion are there between the two given views. This makes the external alignment process (explicit pose estimation and pre-reconstruction) tends to produce unstable results, thus leading to a poor generative NVS performance.

\vspace{-10pt}
\paragraph{Distribution Evaluation.}
In addition to evaluating per-view NVS performance, we assess the distribution of generated novel view sequences using several metrics: Fréchet Inception Distance (FID)~\cite{heusel2017gans}, Fréchet Video Distance (FVD)~\cite{unterthiner2018towards}, and Kernel Video Distance (KVD)~\cite{unterthiner2018towards}. For FID, we treat the novel views as individual images, while for FVD and KVD, we treat them as video clips. We compute these metrics for each model using 1,000 ground truth sequences. To ensure fairness, we report results based on single input view conditions.
Results are shown in~\cref{tab:exp-main-re10k-image-video-distribution}. Our method achieves a comparable FID to ViewCrafter, but outperforms it in both FVD and KVD. This suggests that our method produces more accurate novel views when considering the entire multi-view sequence. Overall, our model generates results that are closer to the ground truth, both in terms of image and video generation.

\begin{table}[t]
  \centering
  \resizebox{0.75\linewidth}{!}{
    \begin{tabular}{c|c c c}
    \toprule
    Method  & FID$\downarrow$ &  FVD$\downarrow$ & KVD$\downarrow$ \\ \midrule
    \quad MotionCtrl~\cite{wang2024motionctrl} \quad & 60.83  & 509.96 & 14.26  \\
     CameraCtrl~\cite{he2024cameractrl} & 52.33 & 561.97 & 24.38 \\
     ViewCrafter~\cite{yu2024viewcrafter} & 46.08 & 485.11 & 13.06 \\
     \methodabbr (Ours) & 46.19 &  425.44  &  8.04\\
    \bottomrule
    \end{tabular}
  }
  \caption{Distribution evaluation on generated views of MotionCtrl~\cite{wang2024motionctrl}, CameraCtrl~\cite{he2024cameractrl}, ViewCrafter~\cite{yu2024viewcrafter}, and our \methodabbr using FID~\cite{heusel2017gans}, FVD~\cite{unterthiner2018towards}, and KVD~\cite{unterthiner2018towards} metrics.}
  \label{tab:exp-main-re10k-image-video-distribution}  
\vspace{-10pt}
\end{table}

\subsubsection{Generative NVS in Objects}
\begin{figure}
  \centering
    \includegraphics[width=1\linewidth]{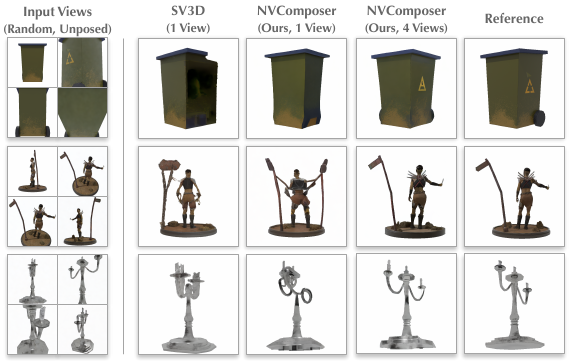}
    \caption{Visual comparison of novel view generation results on the Objaverse~\cite{deitke2023objaverse} test set. All input views are unposed and randomly rendered from the same 3D object.}
    \label{fig:eval-vis-objects}
    \vspace{-15pt}
\end{figure}

In addition to scenes, another important scenario involves generating novel views for synthetic 3D objects. To evaluate the versatility of our proposed pipeline, we compare the generative NVS performance of our model with the object-based generative model SV3D~\cite{voleti2025sv3d} on the Objaverse test set. The numerical and visual results of this comparison are presented in \cref{tab:exp-main-objaverse} and \cref{fig:eval-vis-objects}.
Our model achieves better PSNR and comparable SSIM with SV3D when only a single conditional view is provided. Furthermore, as more unposed input views are added, our model effectively leverages the additional information, producing results that are closer to the real reference.

\begin{table}[t]
  \centering
\resizebox{1\linewidth}{!}{
    \begin{tabular}{c|c|ccc}
\toprule
 Model & Views & PSNR$\uparrow$ & SSIM$\uparrow$ & LPIPS$\downarrow$  \\
\midrule
    SV3D~\cite{voleti2025sv3d} & 1 & 13.8861 & 0.8130 & 0.2731 \\
    \midrule
    \multirow{3}{*}{\methodabbr (Ours)} 
    & 1 & 16.3764 & 0.8218 & 0.2286 \\
    & 2 & 17.1507 & 0.8268 & 0.2067 \\
    & 4 & 17.7234 & 0.8352 & 0.1889 \\
    \bottomrule
    \end{tabular}
    }
  \caption{Generative NVS results on the Objaverse~\cite{deitke2023objaverse} test set. When only a single conditional view is provided, \methodabbr achieves performance comparable to SV3D~\cite{voleti2025sv3d}. As more random unposed condition views are added, \methodabbr’s performance improves significantly.}
  \label{tab:exp-main-objaverse}
\vspace{-20pt}
\end{table}

\begin{table}[t]
  \centering
    \resizebox{0.85\linewidth}{!}{
    \begin{tabular}{c|ccccc}
    \toprule
    Dual-Stream & PSNR$\uparrow$ & SSIM$\uparrow$ & LPIPS$\downarrow$ \\ \midrule
     w/ & 17.0510 & 0.7501 & 0.1353 \\
     w/o  & 14.6857 & 0.7458 & 0.2095 \\ \bottomrule
    \end{tabular}}
  \caption{Ablation experiments on dual-stream diffusion on Objaverse~\cite{deitke2023objaverse}. We train the two models (initialized from the same checkpoint) for one epoch on a small subset of Objaverse. The model without dual-stream only generates images instead of the image-pose bundles.}
  \label{tab:exp-ab-jt}
  \vspace{-5pt}
\end{table}

\begin{table}[t]
  \centering
    \resizebox{0.97\linewidth}{!}{
    \begin{tabular}{c|ccccc}
    \toprule
     Alignment & PSNR$\uparrow$ & SSIM$\uparrow$ & LPIPS$\downarrow$ &  DISTS$\downarrow$\\ \midrule
     w/o & 14.7218 & 0.6291 & 0.3799 & 0.1494 \\
     w/  & 15.6568 & 0.6440 & 0.3284 & 0.1340 \\ \bottomrule
    \end{tabular}
    }
  \caption{Ablation experiments on the geometry-aware feature alignment (\textit{Alignment} in table). We initialize two models with and without the alignment mechanism from a same checkpoint, and train the two models for an epoch, then evaluate them on RealEstate10K~\cite{zhou2018stereo}.}
  \label{tab:exp-ab-fa}
  \vspace{-5pt}
\end{table}

\subsection{Analysis}
\label{sec:analysis}
In this section, we perform several ablation studies and analyses to validate the effectiveness of our model.

\vspace{-10pt}
\paragraph{Ablation on Image-Pose Dual Stream Diffusion.}
To ensure both fairness and feasibility, we train two models with and without the dual-stream diffusion design on a subset of Objaverse~\cite{deitke2023objaverse} containing $5,000$ objects for one epoch from the initial weight of the video diffusion model and evaluate them on a test set with $100$ objects. The results shown in~\cref{tab:exp-ab-jt} demonstrate that the dual-stream diffusion significantly improves the model's performance on generative NVS tasks with unposed multiple condition views.

\begin{figure}[t]
  \centering
    \includegraphics[width=1\linewidth]{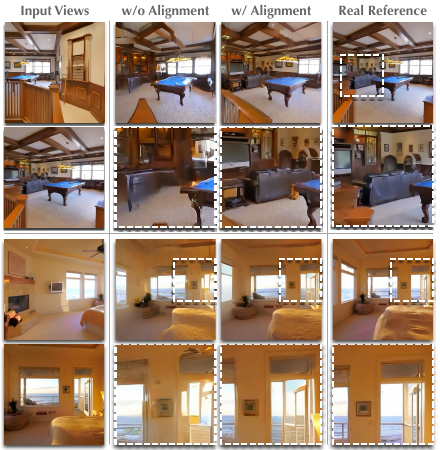}
    \caption{A visual sample in the ablation results of the geometry-aware feature alignment with two input views given. Some patches are zoomed in for a better view. The feature alignment helps \methodabbr to properly utilize contents from other views.}
    \label{fig:exp-ab-fa}
  \vspace{-7pt}
\end{figure}

\vspace{-10pt}
\paragraph{Ablation on Geometry-Aware Feature Alignment.}
We further conduct an ablation study on the geometry-aware feature alignment mechanism using the RealEstate10K dataset~\cite{zhou2018stereo}. In this experiment, we train two models from the same initial checkpoint for one epoch, with and without geometry-aware feature alignment. 
\cref{tab:exp-ab-fa} demonstrate the numerical results and \cref{fig:exp-ab-fa} shows the visualized results of this ablation study. We can clearly tell that this feature alignment mechanism helps our model learn the generative NVS task with unposed multiple conditional views.

\paragraph{Sparse-View Pose Estimation}
Thanks to the unique design of the dual-stream diffusion, \methodabbr can implicitly estimate the pose information. 
We follow the method~\cite{zhang2024cameras} to solve the camera poses from the Plücker rays generated by the dual-stream diffusion of \methodabbr. 
We perform the evaluation on the RealEstate10K dataset, asking the model to estimate the two sparse condition images given in the easy and hard subsets we discussed in \cref{tab:exp-main-re10k}. The accuracy of estimated poses is quantitatively evaluated in the average degrees of rotation angle differences and the average translation difference (with normalization according to the 2-norm of the translation of the furthest view). 
The results are given in ~\cref{tab:exp-pose-est}, where we can see that our method is comparable to the performance of DUSt3R~\cite{wang2024dust3r} in the easy case and outperforms DUSt3R in the hard case.
This is because DUSt3R estimates poses using explicit deep feature correspondences, while \methodabbr implicitly generates pose information during novel view generation. When given sparse condition views with minimal overlap (\emph{i.e.}, in ill-posed cases), our method's implicit pose estimation proves more robust, delivering accurate pose estimates directly corresponding to the current scene in novel view generation.

\begin{table}[ht]
  \centering
  \resizebox{0.85\linewidth}{!}{
    \begin{tabular}{c|c|cc}
    \toprule
    Subset & Method & $\Delta \hat{R}\downarrow$ & $\Delta \hat{T}\downarrow$ \\ \midrule
     \multirow{2}{*}{Easy} & DUSt3R~\cite{wang2024dust3r} & 9.6968 & 0.5757 \\
                           & \methodabbr (Ours)   & 2.7225 & 0.0257 \\
     \midrule
     \multirow{2}{*}{Hard} & DUSt3R~\cite{wang2024dust3r} & 58.3987 &  0.7603 \\
                           & \methodabbr (Ours)   & 5.8566 & 0.0263 \\
    \bottomrule
    \end{tabular}
    }
  \caption{Comparison with pose estimation accuracy on two spare condition images in our RealEstate10K~\cite{zhou2018stereo} test sets. Our \methodabbr implicitly predicts camera poses by generating ray embeddings of condition views while generating target views.}
  \label{tab:exp-pose-est}
  \vspace{-10pt}
\end{table}

\providecommand{\methodabbr}{NVComposer\xspace}
\providecommand{\methodfullname}{Novel View Composer\xspace}
\providecommand{\methodfullnamebold}{\textbf{N}ovel \textbf{V}iew \textbf{Composer}\xspace}

\vspace{-5pt}
\section{Conclusion}
\vspace{-2pt}
We presented \methodabbr, a novel multi-view generative NVS model that eliminates the need for external multi-view alignment, such as explicit camera pose estimation or pre-reconstruction of conditional images. By introducing an image-pose dual-stream diffusion model and a geometry-aware feature alignment module, \methodabbr is able to effectively synthesize novel views from sparse and unposed condition images. Our extensive experiments demonstrate that \methodabbr outperforms state-of-the-art methods that rely on external alignment processes. Notably, we show that the model’s performance improves as the number of unposed conditional images increases, highlighting its ability to implicitly infer spatial relationships and leverage available information from unposed views. This paves the way for more flexible, scalable, and robust generative NVS systems that do not depend on external alignment processes.

{\small \bibliographystyle{ieeenat_fullname} \bibliography{main}}
\end{document}